\documentclass[10pt,twocolumn,letterpaper]{article}

\usepackage{cvpr}              
\usepackage{listings}
\usepackage[dvipsnames,svgnames,table,xcdraw]{xcolor}
\usepackage{amssymb}
\lstdefinestyle{mystyle}{
    backgroundcolor=\color{lightgray!20},   
    commentstyle=\color{green!50!black},    
    keywordstyle=\color{blue},              
    numberstyle=\tiny\color{gray},          
    stringstyle=\color{red!60!black},       
    basicstyle=\ttfamily\footnotesize,       
    breakatwhitespace=false,                 
    breaklines=true,                         
    captionpos=b,                            
    keepspaces=true,                         
    numbers=left,                            
    numbersep=5pt,                           
    showspaces=false,                        
    showstringspaces=false,                  
    showtabs=false,                          
    tabsize=2,                               
    frame=single,                            
    rulecolor=\color{black},                 
}

\usepackage[dvipsnames]{xcolor}

\definecolor{cvprblue}{rgb}{0.21,0.49,0.74}
\usepackage[pagebackref,breaklinks,colorlinks,citecolor=cvprblue]{hyperref}

\def\paperID{*****} 
\def\confName{CVPR}
\def\confYear{2024}

\title{One-Shot Multilingual Font Generation Via ViT
}

\author{
Jiarui Liu\thanks{Equal contribution, listed by last name}\\
University of Massachusetts, Amherst\\
300 Massachusetts Ave, Amherst, MA 01003\\
{\tt\small jiaruil@umass.edu}
\and
Zhiheng Wang\footnotemark[1]\\
University of Massachusetts, Amherst\\
300 Massachusetts Ave, Amherst, MA 01003\\
{\tt\small zhihengwang@umass.edu}
}

\begin{document}
\maketitle
\begin{abstract}
    Font design poses unique challenges for logographic languages like Chinese, Japanese, and Korean (CJK), where thousands of unique characters must be individually crafted. This paper introduces a novel Vision Transformer (ViT)-based model for multi-language font generation, effectively addressing the complexities of both logographic and alphabetic scripts. By leveraging ViT and pretraining with a strong visual pretext task (Masked Autoencoding, MAE) \cite{he_masked_2021}, our model eliminates the need for complex design components in prior frameworks while achieving comprehensive results with enhanced generalizability. Remarkably, it can generate high-quality fonts across multiple languages for unseen, unknown, and even user-crafted characters. Additionally, we integrate a Retrieval-Augmented Guidance (RAG) module to dynamically retrieve and adapt style references, improving scalability and real-world applicability. We evaluated our approach in various font generation tasks, demonstrating its effectiveness, adaptability, and scalability.
\end{abstract}
\section{Introduction}

Designing high-quality fonts is challenging, especially for logographic languages (e.g., Chinese, Japanese, Korean), where thousands of unique characters must be manually created. Existing font generation methods often focus on a single script or rely on large labeled datasets, limiting their ability to handle multiple languages and unseen, custom characters~\cite{azadi_multi-content_2017,li_few-shot_2021,xie_dg-font_2021,he_diff-font_2024}.

To address these challenges, we propose a one-shot multilingual font generation model based on Vision Transformers (ViTs)~\cite{dosovitskiy_image_2021,he_masked_2021}, capable of handling diverse scripts—including Chinese, Japanese, Chinese (CJK) and English—as well as user-invented glyphs. By leveraging ViT-based encoders and decoders, pretrained with a Masked Autoencoder (MAE) objective, our approach eliminates the need for intricate architectural tweaks and robustly captures both global structure and subtle stylistic elements. It supports handwriting and standard fonts without any reference character constraints, outperforming prior methods limited to narrow domains or requiring extensive character libraries~\cite{zhang_separating_2018,lee_arbitrary_2022,sun_learning_2018,xue_metascript_2023}.

Additionally, we integrate a Retrieval-Augmented Guidance (RAG) module~\cite{borgeaud_improving_2022} to dynamically retrieve the most suitable style references from a known inventory, enabling the model to adapt to challenging or unusual characters. Unlike previous approaches that rely on base font references and fail when such references are unavailable, our model eliminates this dependency. It accepts any input shape, including hand-drawn designs, and demonstrates the capability to produce faithful, style-consistent outputs across multiple languages, including unseen fonts, unknown characters, and even user-crafted designs. Experiments show that our ViT-MAE-based system consistently generates high-quality, style-accurate fonts under one-shot conditions, establishing a robust foundation for versatile, cross-lingual font generation.

\section{Related Work} \label{sec}

Font design presents unique challenges, particularly for logographic languages like Chinese and Japanese, which require designing thousands of unique characters individually. While traditional font development remains labor-intensive, deep learning has revolutionized automatic font generation. Early approaches leveraged CNNs and GANs for image-to-image translation, with pioneering works by Azadi \etal \cite{azadi_multi-content_2017} and Fogel \etal \cite{fogel_scrabblegan_2020-1} demonstrating success in alphabetic languages. For logographic languages, models like "Rewrite" \cite{tian_kaonashi-tycrewrite_2024} and "zi2zi" \cite{tian_kaonashi-tyczi2zi_2024} advanced the field through sophisticated style mapping between character pairs.

Recent research has shifted toward few-shot learning using GANs and diffusion models, enabling high-quality font generation from minimal reference characters. Notable advances include Zhang \etal's \cite{zhang_separating_2018} style-content separation networks and GAN-based frameworks by Li \etal \cite{li_few-shot_2021}, Wen \etal \cite{wen_zigan_2021}, and Yu \etal \cite{yu_few_2022}. Hayashi \etal \cite{hayashi_glyphgan_2019} proposed GlyphGAN for ensuring style consistency through independent control of character classes and style vectors. For unsupervised scenarios, Xie \etal's \cite{xie_dg-font_2021} DG-Font maintained structural integrity without extensive paired datasets, while Liu \etal \cite{liu_fonttransformer_2022} introduced FontTransformer for high-resolution synthesis in few-shot settings.

Parallel developments in handwritten text generation have emerged, with Pippi \etal \cite{pippi_handwritten_2023} introducing Transformer-based models for handling unseen styles. MetaScript \cite{xue_metascript_2023} and Tang \etal \cite{tang_few-shot_2022} achieved superior style fidelity in Chinese character generation. Recent diffusion models like FontDiffuser \cite{yang_fontdiffuser_2023} and Diff-Font \cite{he_diff-font_2024} have demonstrated more stable training and higher fidelity for glyph-rich languages.

Masked Autoencoding (MAE) \cite{he_masked_2021} has emerged as a highly effective pretraining strategy for Vision Transformers (ViTs), enabling them to excel in spatial reasoning tasks. By masking a large portion of input image patches and reconstructing the missing information, MAE forces the ViT to learn robust representations of spatial structures and semantic content. This self-supervised approach not only enhances the model’s ability to generalize across diverse visual tasks but also reduces the dependency on extensive labeled datasets. For font generation, where intricate spatial patterns and context relationships are crucial, MAE pretraining empowers the ViT to understand and synthesize complex glyph structures effectively.

Hiera, introduced by Ryali \etal \cite{ryali_hiera_2023}, demonstrates that a simple hierarchical ViT pre-trained with MAE can outperform vision-specific architectures while reducing computational complexity. These approaches highlight the growing utility of ViT-based methods in capturing global context and long-range dependencies~\cite{vaswani_attention_2023, dosovitskiy_image_2021}. By integrating MAE-style pretraining, recent models have effectively bridged the gap between handcrafted design and scalable, data-driven approaches~\cite{he_masked_2021, bao_beit_2022}.

\section{Dataset Description}
Our dataset looks like something as follows: \\
\begin{figure}[h]
    \centering
    \includegraphics[width=0.3\textwidth]{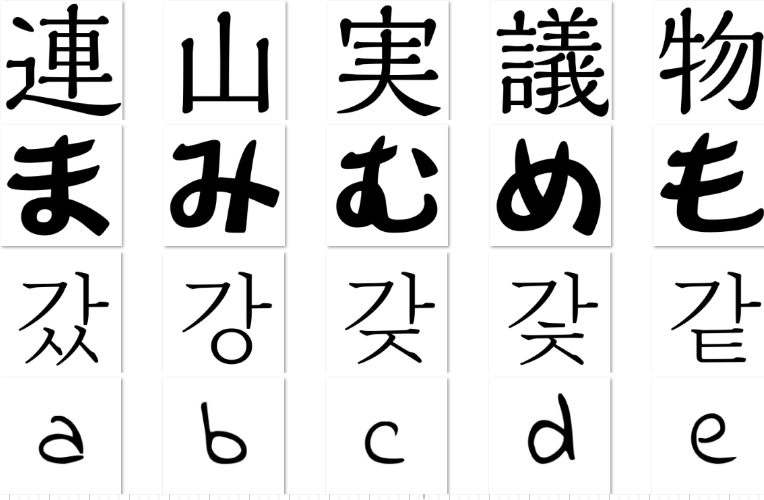}
    \caption{From top to bottom are Chinese, Japanese, Korean and English. The five styles are randomly picked from the total of 308 styles.}
\end{figure}
\\
We have Korean, Chinese, English and Japanese as the four languages that we are using in this model and 308 styles in total. we collected fonts from but not limite to 51 font\cite{noauthor_51font_nodate} and google font\cite{noauthor_browse_nodate}. In the pretraining phase, we skipped Korean. We used total of 154 styles from the other three language with 800k images for pretrain, with the rest of the images split to train validation and test in 8:1:1 ratio, meaning around 1M for training and 150k for validation. As discribed in the dataset section, all the font file have a relative reference character. In our training phase, we decide to let the input image has or don't have the access to the reference character. Thus, to test our model throughout, we decided to distributes it randomly to four sets: content font unseen (20k), content reference character unseen (50k), style reference character unseen (35k) and style font unseen (80k).
\section{Method}
\subsection{Dataset Preparation and Pretraining}
The default VitMAE model is able to take 80*80 images, however, our initial training shows that the model does not learn well from such big size. Therefore, we reduce the size of the model to 24*24, and used the patch size of 16.\\
Our goal is to pretrain the model with part of our dataset to let the model learn the encoder and decoder that we are going to use in the later training. This process also significantly reduced our learning time, and it is proven to be capable to reconstruct the font picture and learn its pattern which we will benefit when used for our encoder and decoder in the main model.\\
The original Vit-Mae model have a masked ratio of 0.75, after some experiment, we realized that this mask ratio might be too high for some character font pictures because of its complexity, thus we reduced the mask rate to 0.65. Below are the two pictures comparing the two mask ratio.\\

\begin{figure}[htbp]
    \centering
    \begin{subfigure}[b]{0.3\textwidth}
        \centering
        \includegraphics[width=\textwidth]{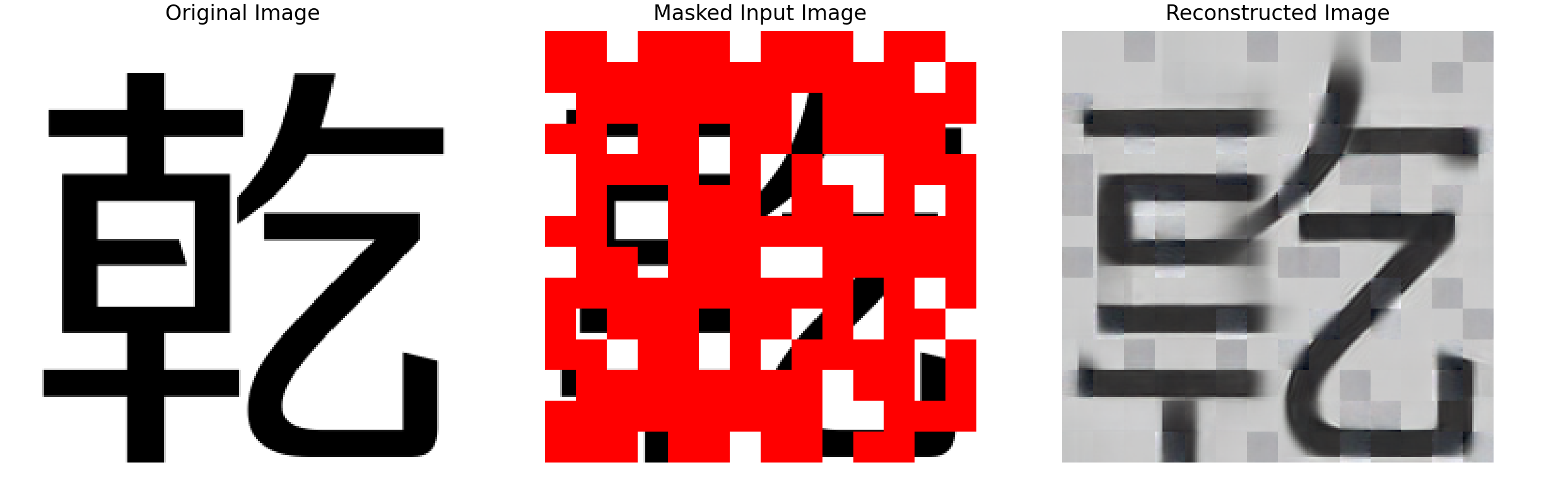}
        \caption{The reconstruction result using the Vit-MAE-base model with the 0.75 mask ratio, we can see that the a lot of strokes are not properly reconstructed}
        \label{fig:first}
    \end{subfigure}
    \hfill
    \begin{subfigure}[b]{0.3\textwidth}
        \centering
        \includegraphics[width=\textwidth]{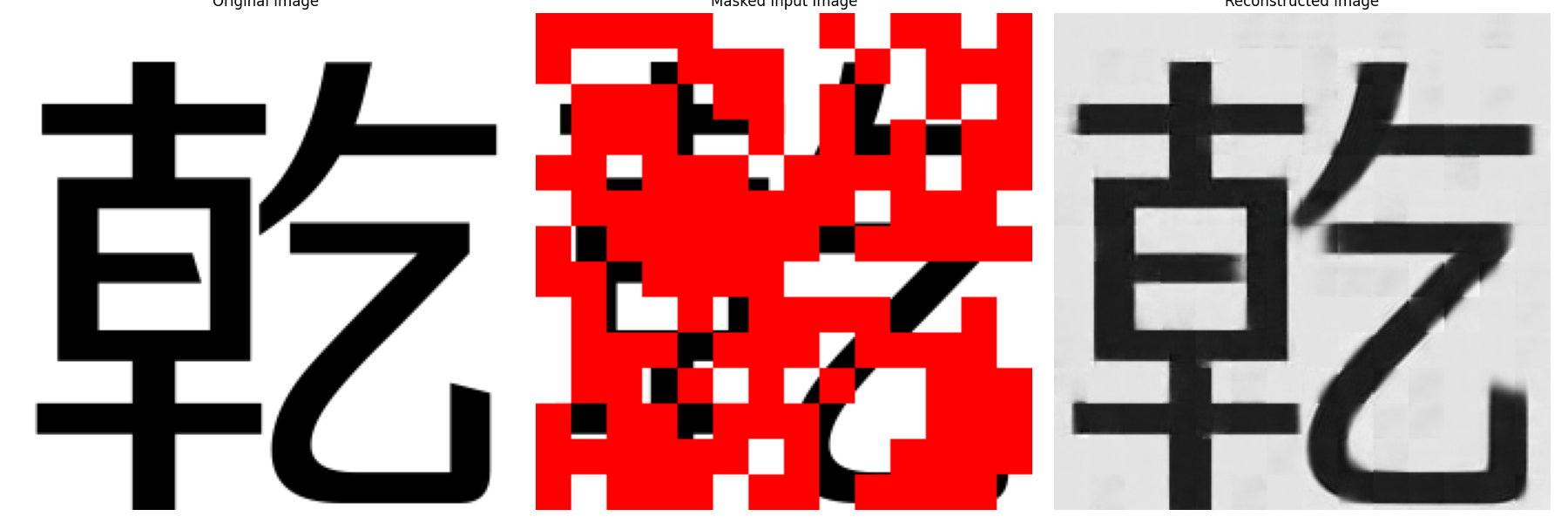}
        \caption{This is the reconstruction result using the modified mask ratio of 0.65, after 17 epochs, we can see that the image is pretty well reconstructed.}
        \label{fig:second}
    \end{subfigure}
    \caption{The comparison showed that it is necessary to pretrain the VitMAE model in our dataset. This will help us to start with a confident encoder and decoder for the main model.}
    \label{fig:comparison}
\end{figure}

\subsection{General Overview}
Vision Transformers (ViTs) are well-suited for font image generation due to their ability to capture both global structures and long-range dependencies. Since each glyph can be considered a single-object input of fixed position and size, the self-attention mechanisms inherent in ViTs enable simultaneous modeling of content and style features.

Our approach builds upon a previous bi-encoder architecture by introducing a transformer-based generator guided by content and style embeddings. Specifically, we employ two ViT-based encoders: a content encoder that extracts the structural features of an input glyph, and a style encoder that summarizes visual style cues from multiple reference images. By combining these representations, our generator produces novel glyphs that retain the original content while reflecting the target style.

The training objective involves minimizing both reconstruction errors and perceptual differences between the generated output and the ground truth. In addition to standard pixel-wise losses (e.g., L2), we incorporate perceptual metrics derived from a pre-trained VGG19 network. This encourages alignment with high-level visual features. While adversarial methods could further enhance realism, our primary focus is on producing accurate, style-consistent reconstructions guided by explicit content and style features.

Building on prior work, our goals are:
\begin{enumerate} 
    \item \textbf{Enhanced Style and Content Feature Extraction:} We refine our architecture and training strategies to improve the efficiency and fidelity of feature extraction, ensuring that the model effectively captures both the structural details and stylistic nuances of glyphs.
    \item \textbf{Improved Multi-Lingual Performance:} Since fonts often must support a variety of scripts, we adapt our model to handle multi-lingual datasets. This ensures that it generalizes well to diverse glyph shapes and stylistic variations.
\end{enumerate}

\subsection{Baseline}
To benchmark our advancements, we implemented four baseline models that share a general structure with our main approach but differ in components:

\begin{enumerate}
    \item \textbf{ViT from Scratch (Grayscale) + MSE:} A ViT encoder-decoder from scratch trained on grayscale inputs using only MSE loss.
    \item \textbf{ViT from Scratch (RGB) + MSE:} Similar to the above, but trained on RGB inputs to utilize color information. We thought this one will be more computational expensive than the gray scale version, but it turns out they are nearly the same.
    \item \textbf{Facebook-ViTMAE-base + MSE:} A ViTMAE-base model (from Facebook) serving as encoder and decoder, with MSE loss guiding reconstruction.
    \item \textbf{Pretrained ViTMAE + MSE:} A ViTMAE model pretrained on our dataset, retaining MSE as the loss function.
\end{enumerate}

With the baseline and the main model trained on the same dataset, the baselines provide reference points against which we can measure the improvements introduced by our combined-loss, cross-attention bi-encoder method. 

\subsection{Main Method -- Cross-Attention Bi-Encoder With Combined Loss}

Our core architecture integrates a cross-attention mechanism to fuse content and style representations effectively. This approach diverges from simpler additive or concatenation-based methods by allowing the content embedding (queries) to selectively attend to and incorporate the most relevant stylistic elements (keys and values) from the style embedding. Through this targeted interaction, we ensure that the structural fidelity of the glyph is preserved while layering on the nuanced aesthetic traits of the desired style.

The model is built upon a Vision Transformer MAE (Masked Autoencoder) backbone and comprises three principal components:

\begin{enumerate}
    \item \textbf{Content Encoder:}  
    A ViT-based encoder that processes the input glyph image, producing a content embedding representing the character’s structure and global shape.

    \item \textbf{Style Encoder:}  
    Another ViT-based encoder, similarly initialized, processes reference style images to produce an embedding capturing stylistic attributes such as stroke thickness, curvature, and texture patterns.

    \item \textbf{Cross-Attention Module and Decoder:}  
    The cross-attention module uses the content embedding as queries and the style embedding as keys/values, ensuring that style information is integrated into the content representation in a controlled, attention-driven manner. A ViT-based decoder, adapted from the MAE framework, reconstructs the complete glyph from these fused embeddings, compensating for initially masked patches. This decoder leverages the learned representations to generate a final image that mirrors the style’s character while maintaining correct glyph shape.
\end{enumerate}

\begin{figure}[h]
    \centering
    \includegraphics[width=0.4\textwidth]{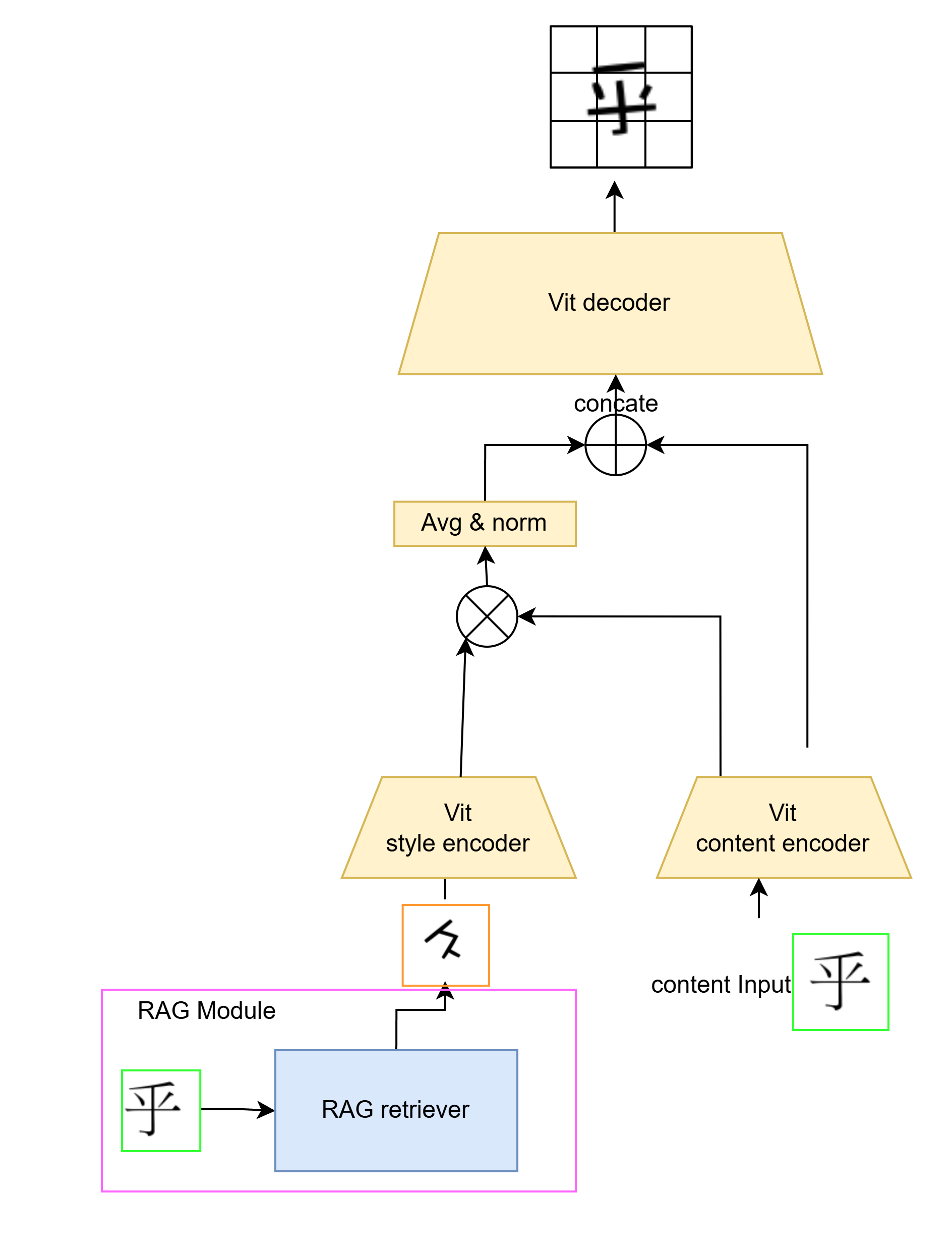}
    \caption{Our proposed model utilizes a cross-attention mechanism to guide the fusion of content and style embeddings, enhancing the flexibility and fidelity of glyph generation. Noted, the pink boxed RAG module is an add on to our main model. The green boxed font images are content input, and the orange boxed font image is style input.}
\end{figure}

\subsubsection*{Choice of Perceptual Layers for Loss Computation}

Our combined loss function uses perceptual losses derived from a pre-trained VGG19 network. We select specific layers to extract both content and style features, ensuring the model captures the structural integrity and stylistic nuances of the target glyph:

\begin{itemize}
    \item \textbf{Content Layers (e.g., `$relu2_2$`):}  
    Mid-level layers of VGG19 represent more abstract features than the very early layers (which primarily capture edges and simple textures) but are not as abstract as the deepest layers. For fonts, these mid-level layers are well-suited to capture the shape and arrangement of strokes and overall glyph structure, ensuring the output maintains the identity of the character.

    \item \textbf{Style Layers (e.g., `$relu1_1$`, `$relu2_1$`, `$relu3_1$`, `$relu4_1$`, `$relu5_1$`):}  
    Lower-level layers focus on simple textures and patterns, while deeper layers capture increasingly complex structures. By sampling style features from multiple layers spanning early to deeper stages, we glean information about fine-grained textures, subtle stroke thickness variations, and higher-level aesthetic patterns. This multi-layer approach ensures that style loss guides the model to produce glyphs that faithfully reflect the chosen style across a wide range of visual attributes.
\end{itemize}

\subsubsection*{Combined Loss Formulation and Rationale}

We define a combined loss that balances content fidelity, stylistic accuracy, and pixel-level precision:

\begin{equation}
\mathcal{L}_{total} = \alpha \mathcal{L}_{content} + \beta \mathcal{L}_{style} + \gamma \mathcal{L}_{MSE}
\end{equation}

Here, $\mathcal{L}_{content}$ and $\mathcal{L}_{style}$ are the aforementioned perceptual losses derived from selected VGG19 layers, while $\mathcal{L}_{MSE}$ enforces pixel-level alignment.

From initial experiments, we observed that the raw scales of these losses differed substantially. If all losses were weighted equally, the MSE term risked becoming negligible, and the style cues might not be fully expressed. Since our primary objective is to perform style transfer, we emphasize $\mathcal{L}_{style}$ by setting a relatively higher weight $\beta = 0.4$ compared to $\alpha = 0.1$ for content. This ensures that the model focuses more on recreating the stylistic aspects accurately. At the same time, to maintain relevance for pixel-level details, we increase $\gamma$ to $1.0$, ensuring that the MSE term provides a stable, fine-grained anchor for sharpening edges and ensuring clear glyph boundaries.

\subsubsection*{Post-Training Refinement with L1 Loss}

Following approximately 10 epochs of training with the combined loss, we introduce an additional short (0.5 epoch) refinement phase using an L1 loss in place of the perceptual losses:

\begin{equation}
\mathcal{L}_{refine} = \| I_{pred} - I_{gt} \|_1
\end{equation}

The motivation here is to smooth out residual artifacts, reduce subtle noise, and clean up the generated glyph images. L1 loss encourages sparsity in the error, which helps refine fine details and ensures the final output is both visually coherent and stylistically faithful.

\subsubsection*{Retrieval-Augmented Guidance (RAG) Module}

To further enhance adaptability, we integrate a Retrieval-Augmented Guidance (RAG) module for scenarios where the user already has a set of style references available. In this scenario, the task input consists of a target content character and a desired style. The RAG module identifies the most suitable style reference from the known set and provides it to the model to generate the desired output.

The central hypothesis is that each character embodies a subset of the font style, and certain style features may not be fully represented in the input style reference characters. Characters with similar content or structure can provide additional style information during style transfer. \textbf{We hypothesize that the retrieval-enhanced method will extract the most relevant style information from the available reference set, enabling the transformer model to learn and replicate the style more precisely and efficiently.}

For each style, we use the content encoder to create embeddings for each character image and employ FAISS to build an exact search index.

\subsubsection*{Key Steps of the RAG Module}
\begin{enumerate}
    \item \textbf{Embedding Extraction:}  
    For each available font style, we use our model's content encoder to generate content representations, which are concatenated to become the embedding. The embedding dimensions is \texttt{hidden\_size $\times$ (patch\_num + 1)}.

    \item \textbf{FAISS Indexing:}  
    We construct an exact search index for each style using FAISS. The \textit{IndexFlatL2} index is employed to perform similarity searches based on cosine similarity.

    \item \textbf{Nearest Neighbor Retrieval:}  
    Given a new content glyph and a target style, we compute the glyph’s embedding and retrieve the most similar style references using nearest-neighbor search with our built embeddings.

    \item \textbf{Font Character Generation:}  
    The selected style reference exemplar is passed to the model, providing additional context to generate the target character.
\end{enumerate}

The RAG module enhances scalability and flexibility, while also addressing some hard failure cases by dynamically retrieving style references that complement the content glyph. This approach, combined with our cross-attention bi-encoder ViT model and refined loss strategy, ensures a more robust and versatile glyph generation framework.

\section{Benchmark Example -- DiffusionFont}
Among current state-of-the-art style transfer models, DiffuserFont is the only input-compatible method that we can feasibly test against. Although its focus is on Chinese characters, comparing our results with DiffuserFont provides a useful qualitative benchmark.

\begin{figure}[h]
    \centering
    \includegraphics[width=0.3\textwidth]{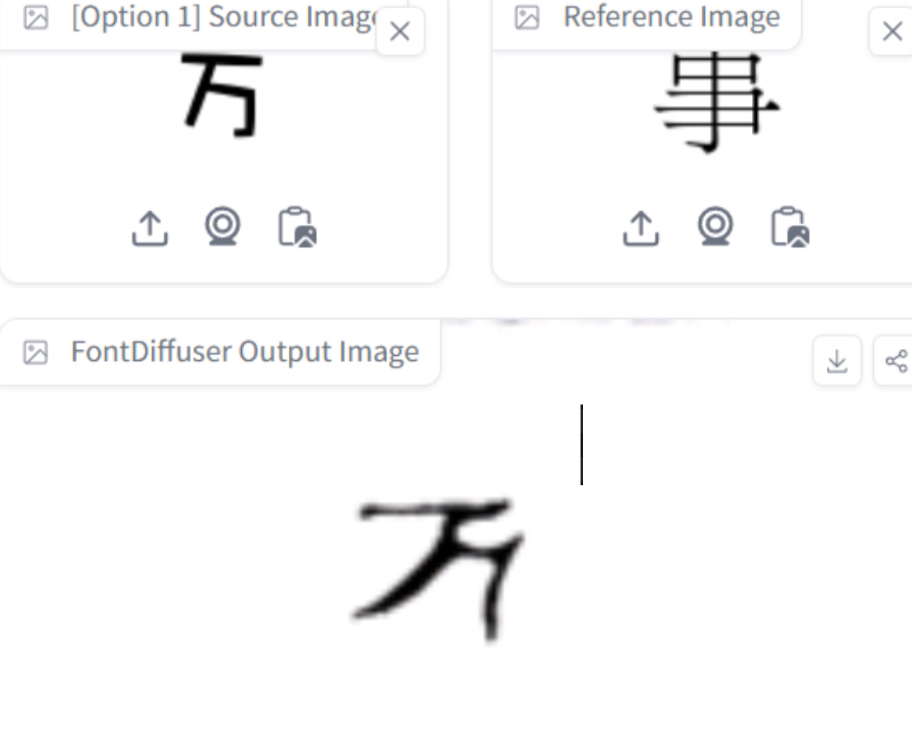}
    \caption{DiffuserFont captures content but deviates significantly in style.}
    \label{fig:example_image}
\end{figure}

While we illustrate with Chinese characters here, later evaluations consider all four languages in our dataset.
\section{Experiment}
As discussed in the Methods section, our experimental setup involves providing the model with a content image, a style image, and a corresponding ground truth image. The content and ground truth share the same underlying character, whereas the style and ground truth share the same stylistic attributes. Our approach utilizes content and style encoders derived from a pretrained ViTMAE model, enabling the network to efficiently capture content and style features. The different loss functions are then applied to balance the content fidelity, style accuracy, and pixel-level alignment, while the decoder component, also adapted from the pre-trained ViTMAE model, aims to reconstruct the target glyph. 

\subsection{Baseline models}
As our baseline model uses the MSE difference between the ground truth and the generated image for the loss function, we believe that this is one of the naive ways to compare the difference between the two images. However, when we monitored the loss log, we realized that it takes around 3.5 hours to train on four 4090 per epoch and the loss goes down very slow. The sample results are shown as follows:
\begin{figure}[h]
    \centering
    \includegraphics[width=0.5\textwidth]{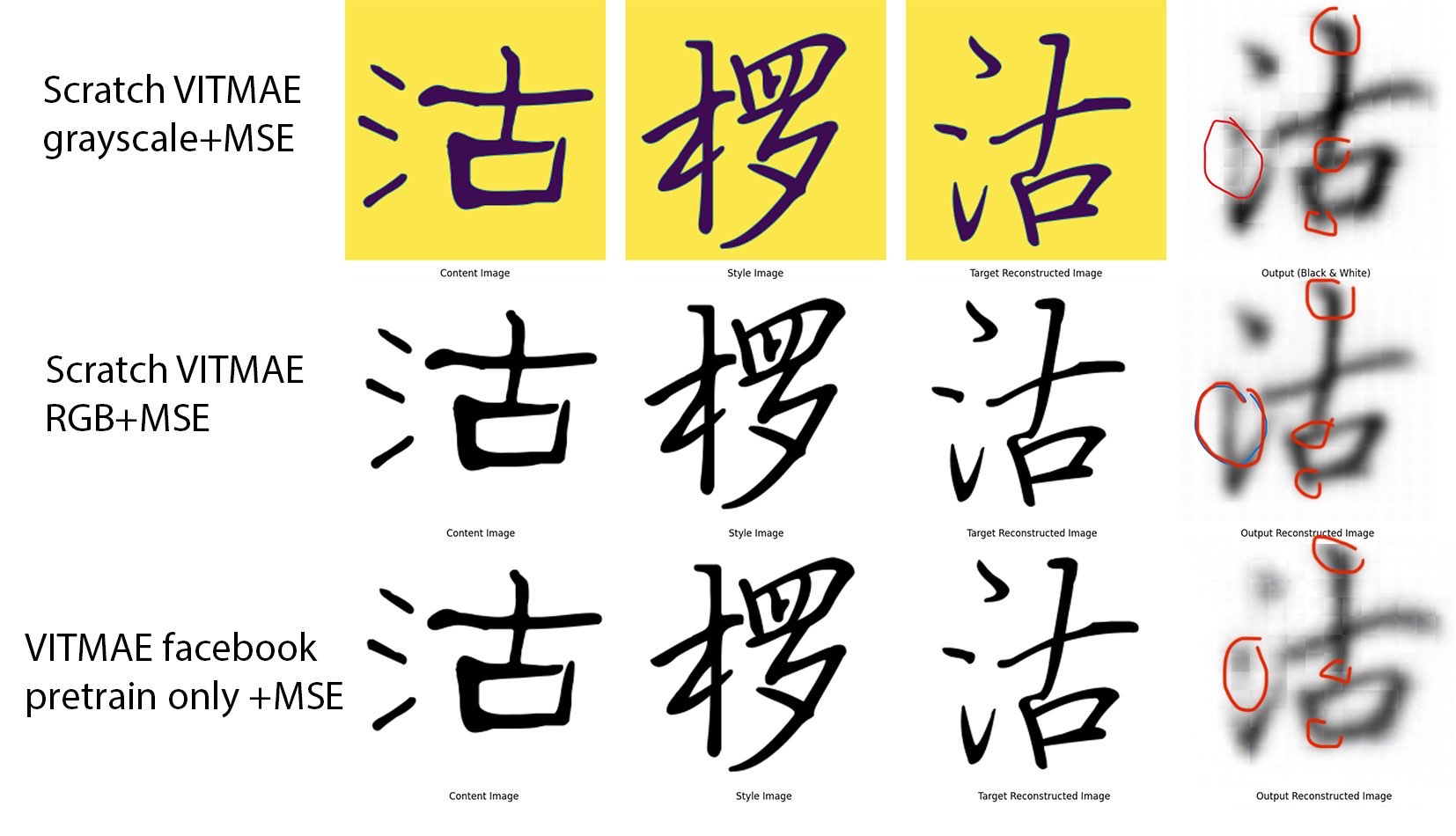}
    \caption{In each row, the first image is the content image, the second image is the style image, the third image is the ground truth image, and the last image is the generated image. The example is randomly picked from four testing set.}
\end{figure}
\\
We used the same content input and style input for the three models to better compare it. All three models are trained with 10 epochs. At first glance, we can see that the stroke widths differ significantly from those in the ground truth. By examining the circled sections and comparing them to the corresponding ground truth images, it becomes clear that these models fail to capture the detailed stylistic elements we intended them to learn. However, we can see that the content is well preserved from the content input, thus MSEs are able to learn the content, but they do not perform well for transferring styles.

\subsection{Bi-encoder Model}
We run the same dataset for pretraining, training, validation, and testing across all models. The test results generated in the following are five randomly selected images from the four testing set. Unfortuantely, as english only have 26 characters and 11 styles, we did not have it covered in the chart, but they are computed in Table 1 and Table 2 below.\\
\subsubsection{Human Evaluation on Single Language Style Transfer}
\begin{figure}[h]
    \centering
    \includegraphics[width=0.38\textwidth]{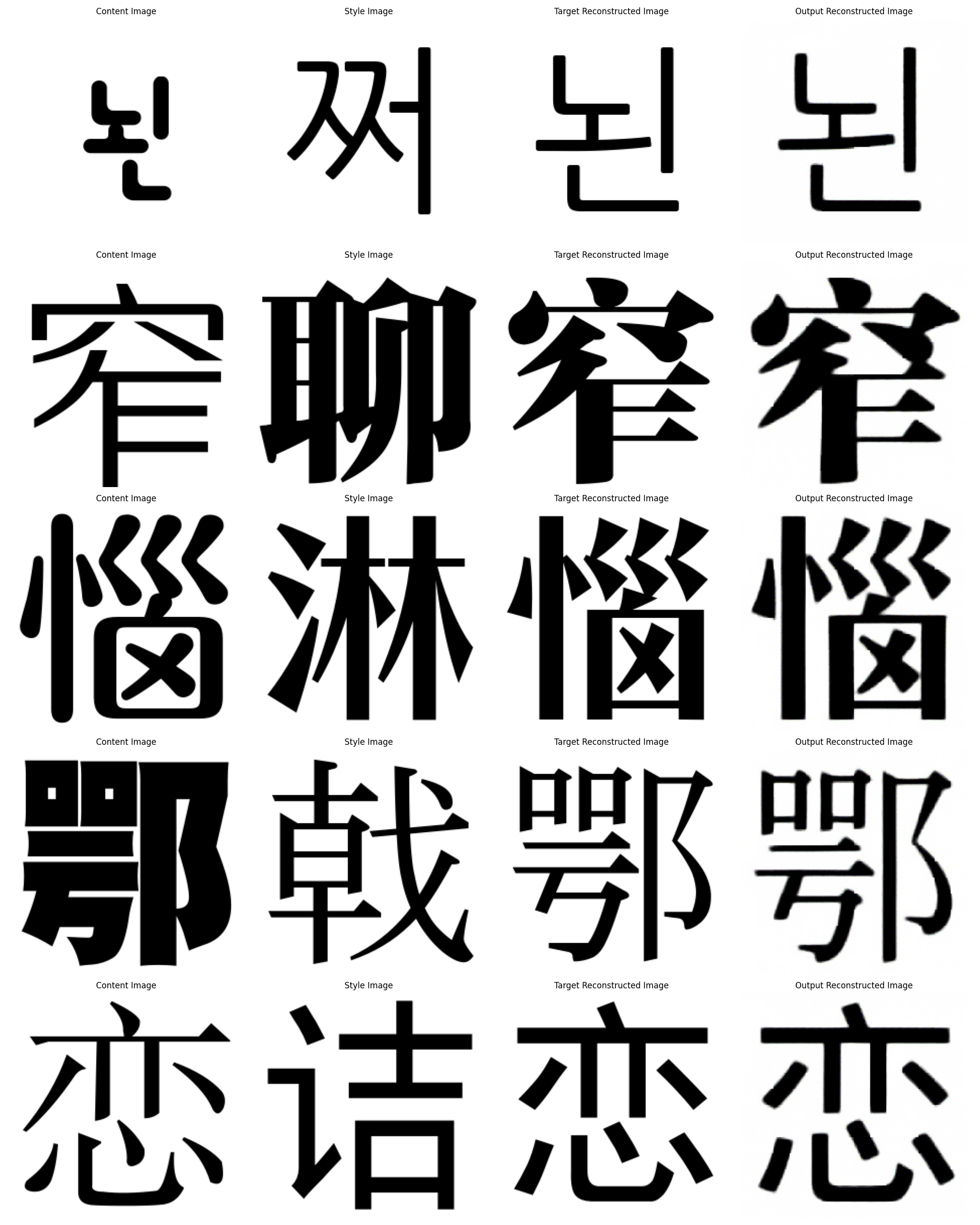}
    \caption{During the human evaluation, we covered the third column for the subject to determine the success of the style transfer. We have shown them 20 comparison in total, this is only an example. Because all of the subjects speaks english, so the enlish style transfer is not used for human evaluation}
\end{figure}

We recruited six subjects, divided into two groups. Group 1 (three subjects) spoke Chinese, Japanese, or Korean, and Group 2 (three subjects) spoke only English. Each subject rated style transfer on a scale of 0 (no transfer) to 2 (complete transfer).

All three participants in Group 1 rated the transfer as 2, indicating clear recognition of the intended style. In Group 2, two participants rated it as 2, while one gave a 1, noting difficulty in assessing small strokes. Group 1 members mentioned that viewing the characters from a distance helped confirm successful style transfer, while the Group 2 participant who scored it lower struggled with subtle details.

\subsubsection{Cross-Language Style Transfer}

\begin{figure}[h]
    \centering
    \includegraphics[width=0.5\textwidth]{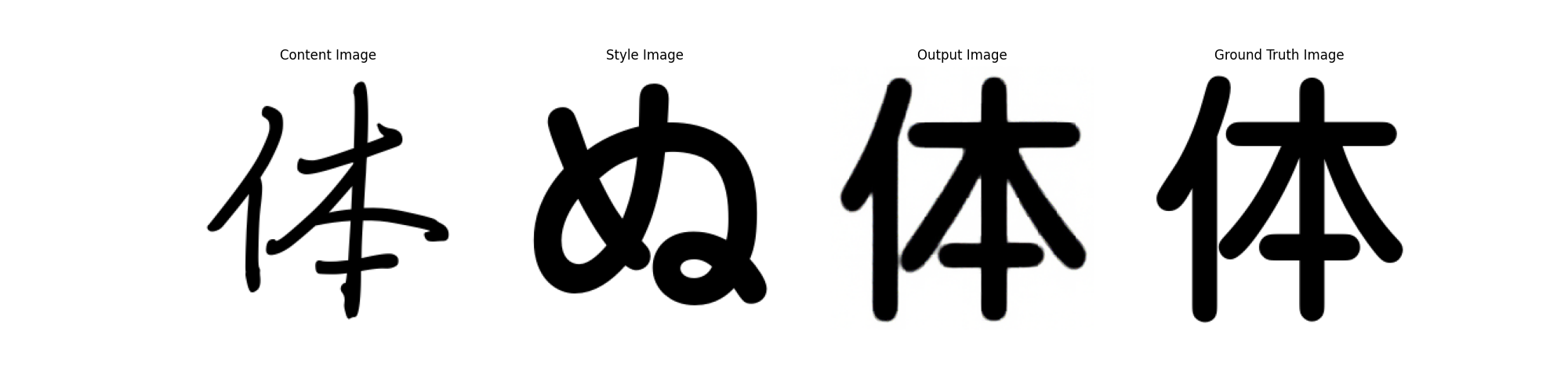}
    \caption{The input content is Chinese, and the style input is Japanese hiragana. The generated result closely matches the ground truth, demonstrating our model's capability for cross-language style transfer.}
    \label{fig:china_to_hiragana}
\end{figure}

\begin{figure}[htbp]
    \centering
    \begin{subfigure}[b]{0.4\textwidth}
        \includegraphics[width=\textwidth]{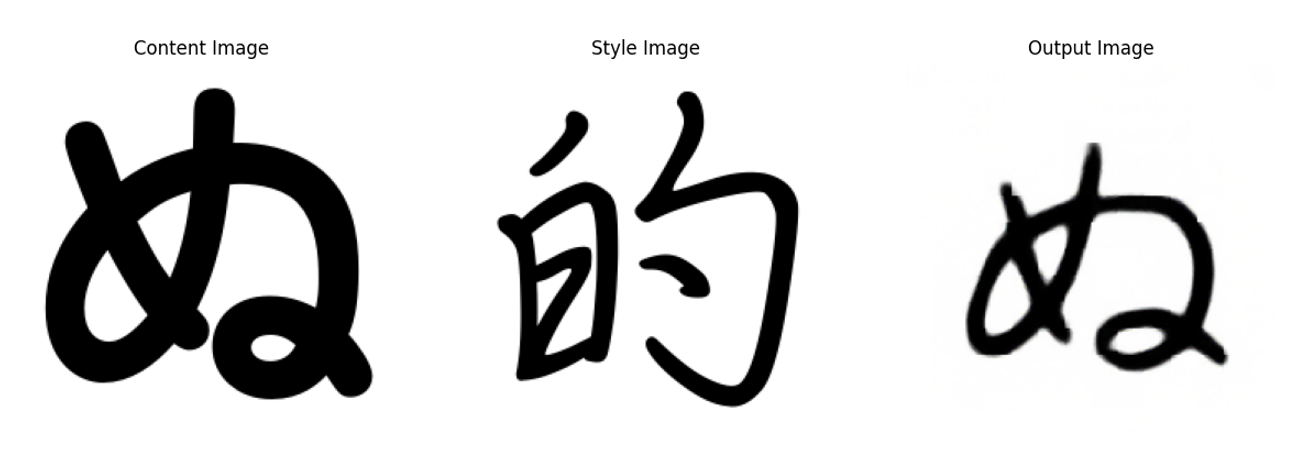}
        \caption{}
        \label{fig:hiragana_chinese_style_a}
    \end{subfigure}
    \hfill
    \begin{subfigure}[b]{0.4\textwidth}
        \includegraphics[width=\textwidth]{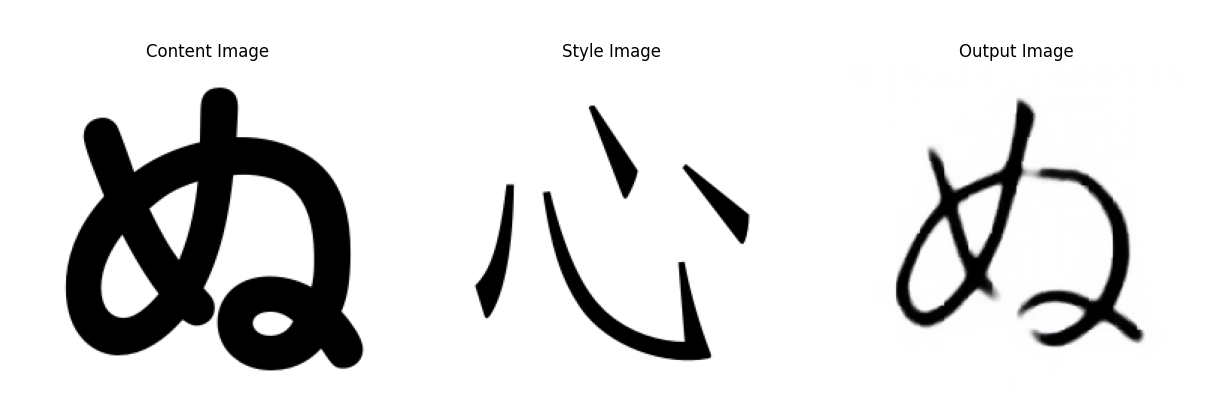}
        \caption{}
        \label{fig:hiragana_chinese_style_b}
    \end{subfigure}
    \caption{Subfigures (a) and (b) show the same Japanese hiragana content paired with different Chinese style inputs. Both examples involve unseen content, unseen style, and no direct character reference.}
    \label{fig:hiragana_chinese_style}
\end{figure}

Comparing Figures~\ref{fig:china_to_hiragana} and \ref{fig:hiragana_chinese_style}, we see that the bi-directional setup enables effective cross-language training. Furthermore, our model exhibits robustness beyond that of existing font transfer methods: it does not require a character reference for either the style or content input. In contrast, other approaches cannot produce any output without a referenced character in their library. In the next section, we further demonstrate the effectiveness of our model.

\subsubsection{Made-Up Content, Unseen Styles, and Handwriting}\label{Made-Up}
\begin{figure}[h]
    \centering
    \includegraphics[width=0.5\textwidth]{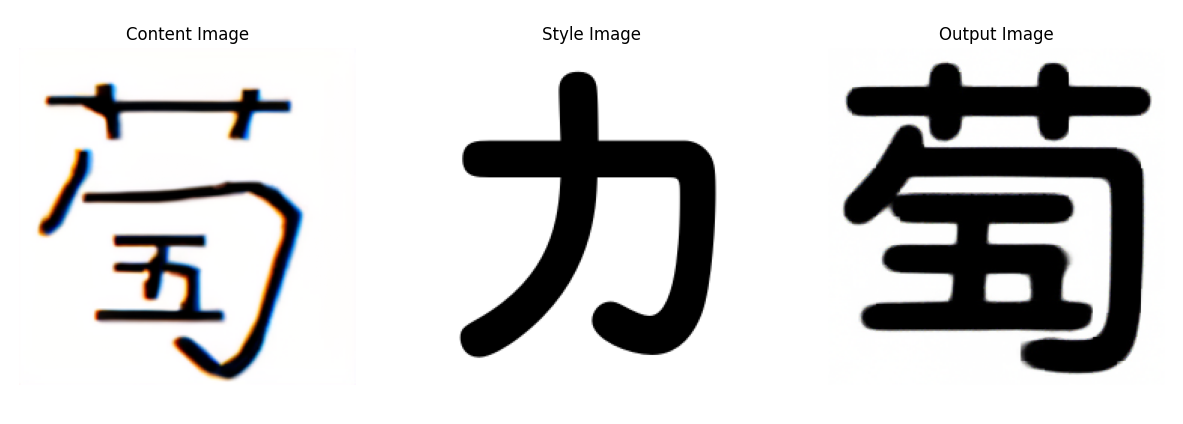}
    \caption{The input content is a made-up handwritten word, and the style input is also an unseen style. As a result, there is no ground truth reference for this example.}
    \label{fig:made_up_handwriting}
\end{figure}

Conventional approaches typically focus solely on either handwriting style transfer or standardized font transfer and often require a reference character in their library to produce any output. Our model, however, handles both scenarios seamlessly—even when no reference character is available. As shown in Figure~\ref{fig:made_up_handwriting}, our method successfully interprets a purely handwritten, invented input and applies an unseen style, demonstrating its adaptability and robustness.

\subsubsection{Numerical Comparison}

Table~\ref{tab:test_settings} shows that our method attains relatively strong performance under challenging conditions. In contrast, as seen in Table~\ref{tab:comparison}, both DF-Font and CF-Font were trained for significantly longer (20k iterations vs. our 10), used a few-shot rather than one-shot approach, and benefited from more thoroughly tuned hyperparameters.
It is also critical to note that our evaluation dataset is more difficult than those used in prior work. Moreover, our model, trained on a dataset of approximately 2.5 million samples across 308 styles, is substantially larger than the datasets employed by DF-Font and CF-Font—ours being more than 50 times their size. Their training setup required about 15 hours on four V100 GPUs for full convergence, while our setup, after an equivalent training duration on four RTX 4090 GPUs, reached only about 10 iterations due to the substantially larger dataset. Thus, if given more training time and computational resources, we expect our model’s performance to surpass the current results. Furthermore, their models cannot generate characters not present in their reference libraries, whereas ours excels in generalization and scalability. Despite these disadvantages, our model’s results are not far behind the current state of the art, and we are confident that with additional training iterations, we will achieve even better outcomes.

We have not included Diffusion-Font in our comparison. Although it is also a one-shot method and reports superior metrics compared to CF-Font and DG-Font in their paper, our empirical tests indicate that Diffusion-Font cannot consistently perform valid style transfer. Given these observed shortcomings, we remain skeptical about their reported results.

\begin{table*}[t]
\centering
\caption{Performance metrics across different unseen font scenarios. 
\textbf{SS} refers to Style Font Unseen, \textbf{SC} refers to Style Reference Character Unseen, \textbf{CS} refers to Content Font Unseen, and \textbf{CC} refers to Content Reference Character Unseen. 
The dataset consists of all four languages.
Our method demonstrates robust performance in font generation across multiple languages without requiring knowledge of input character content or style, highlighting its generalization.}
\begin{tabular}{cccc|ccccc}
\toprule
\multicolumn{4}{c|}{\textbf{Unseen Settings}} & \multicolumn{5}{c}{\textbf{Metrics}} \\
\cmidrule(lr){1-4} \cmidrule(lr){5-9}
\textbf{SS} & \textbf{SC} & \textbf{CS} & \textbf{CC} & \textbf{L1 Loss ↓} & \textbf{MSE ↓} & \textbf{SSIM ↑} & \textbf{LPIPS ↓} & \textbf{FID ↓} \\
\midrule
$\checkmark$ & $\times$ & $\times$ & $\times$ & 0.18697 & 0.54534 & 0.66566 & 0.19229 & 25.56092 \\
$\checkmark$ & $\checkmark$ & $\times$ & $\times$ & 0.18952 & 0.54987 & 0.66362 & 0.19494 & 25.23063 \\
$\times$ & $\times$ & $\checkmark$ & $\times$ & 0.18781 & 0.54672 & 0.66534 & 0.19200 & 25.52842 \\
$\times$ & $\times$ & $\checkmark$ & $\checkmark$ & 0.18879 & 0.54857 & 0.66485 & 0.19439 & 26.15635 \\
\bottomrule
\end{tabular}
\label{tab:test_settings}
\end{table*}

\subsubsection{RAG Module}
\begin{table*}[t]
\centering
\caption{Performance metrics across different unseen font scenarios for our model enhanced with the RAG module. Surprisingly, RAG does not improve standard evaluation metrics, including L1 Loss, RMSE, SSIM, LPIPS, and FID.}
\begin{tabular}{cccc|ccccc}
\toprule
\multicolumn{4}{c|}{\textbf{Unseen Settings}} & \multicolumn{5}{c}{\textbf{Metrics}} \\
\cmidrule(lr){1-4} \cmidrule(lr){5-9}
SS & SC & CS & CC & \textbf{L1 Loss ↓} & \textbf{MSE ↓} & \textbf{SSIM ↑} & \textbf{LPIPS ↓} & \textbf{FID ↓} \\
\midrule
$\checkmark$ & $\times$ & $\times$ & $\times$ & 0.19858 & 0.56531 & 0.65361 & 0.20315 & 26.49332 \\
$\checkmark$ & $\checkmark$ & $\times$ & $\times$ & 0.19875 & 0.56551 & 0.65507 & 0.20319 & 26.37470 \\
$\times$ & $\times$ & $\checkmark$ & $\times$ & 0.19812 & 0.56434 & 0.65462 & 0.20121 & 26.61482 \\
$\times$ & $\times$ & $\checkmark$ & $\checkmark$ & 0.19930 & 0.56612 & 0.65337 & 0.20307 & 27.13604 \\
\bottomrule
\end{tabular}
\label{tab:rag_test_settings}
\end{table*}

The evaluation results of our model with the Retrieval-Augmented Guidance (RAG) module (Table~\ref{tab:rag_test_settings}) indicate that the integration of RAG does not yield improvements across standard evaluation metrics, including L1 Loss, RMSE, SSIM, LPIPS, and FID, in unseen font scenarios. However, subsequent human analysis revealed that RAG effectively addresses specific failure cases and improves performance on hard examples, thereby enhancing the model's usability and adaptability. This demonstrates that while RAG does not directly impact traditional quantitative metrics, it plays a vital role in practical applications by ensuring robustness and reliability in challenging scenarios.

\begin{figure}[h]
    \centering
    \includegraphics[width=0.4\textwidth]{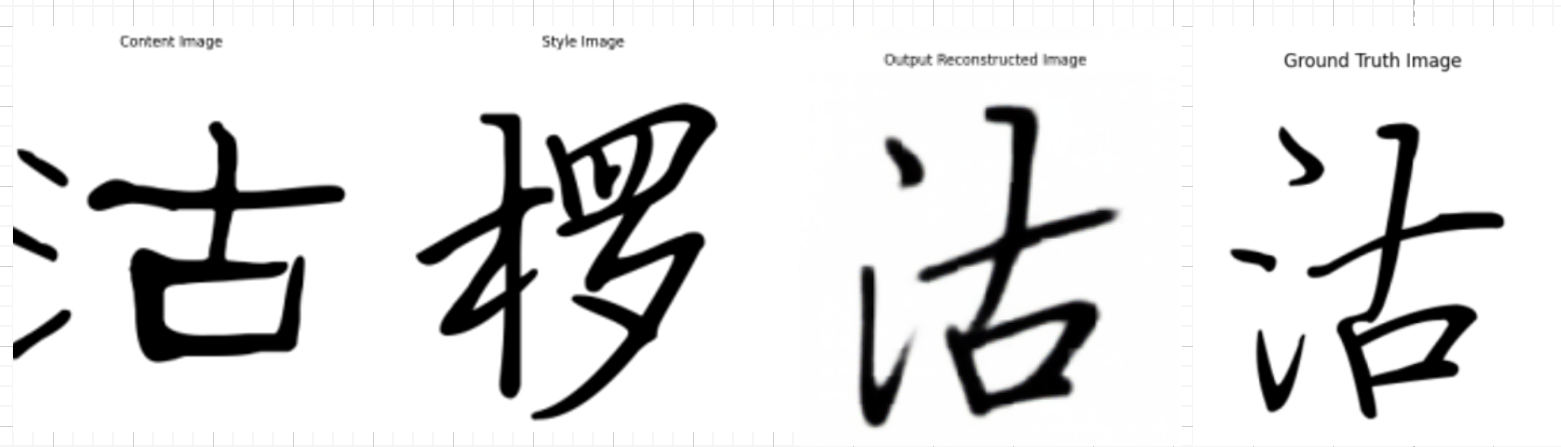}
    \caption{The target character includes the three-dot "water" radical, but the model initially generates the two-dot "ice" radical. The RAG module successfully retrieves the correct three-dot radical and enables the model to produce the desired glyph.}
    \label{fig:fail_case2}
\end{figure}

\begin{figure}[h]
    \centering
    \includegraphics[width=0.3\textwidth]{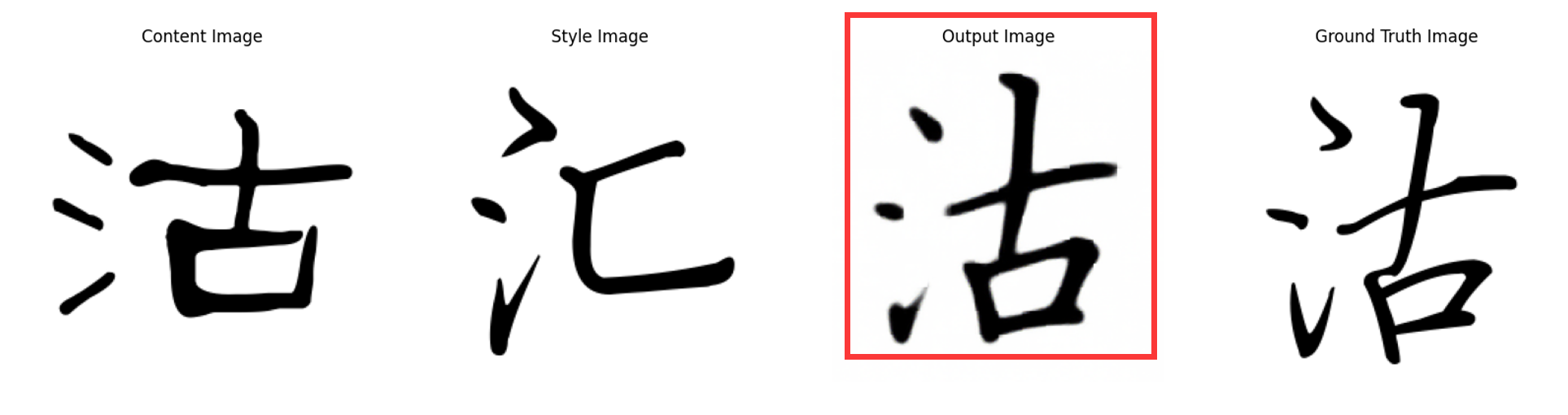}
    \caption{The RAG module resolves errors in generating the three-dot "water" radical by retrieving style information from a known set of characters.}
\end{figure}

\begin{figure}[h]
    \centering
    \includegraphics[width=0.3\textwidth]{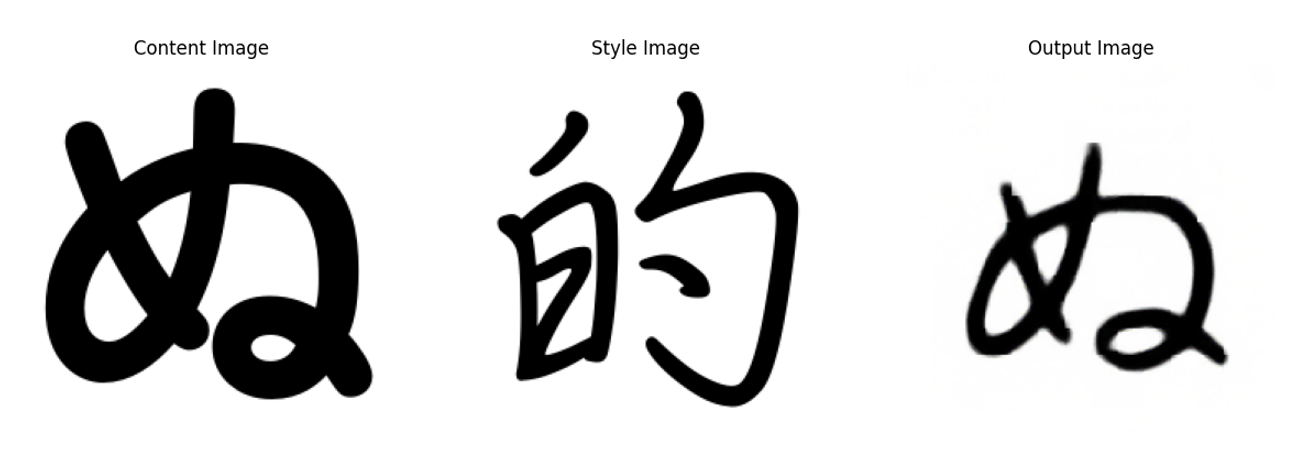}
    \caption{For the challenging task of generating hiragana characters in a Chinese handwriting style, the RAG module retrieves structurally similar Chinese characters (historically related to the target hiragana) and produces highly accurate results.}
    \label{fig:hiragana_use_char_example}
\end{figure}

\begin{table*}[t]
\centering
\caption{Performance metrics for unseen font style (SS). 
The evaluation results for DG-Font and CF-Font are sourced from the CF-Font paper. }
\begin{tabular}{ccccc|ccccc}
\toprule
\multicolumn{5}{c|}{\textbf{Unseen Settings}} & \multicolumn{5}{c}{\textbf{Metrics}} \\
\cmidrule(lr){1-5} \cmidrule(lr){6-10}
SS & SC & CS & CC &  & \textbf{L1 Loss ↓} & \textbf{RMSE ↓} & \textbf{SSIM ↑} & \textbf{LPIPS ↓} & \textbf{FID ↓} \\
\midrule
$\checkmark$ & $\times$ & $\times$ & $\times$ & DG-Font & 0.07841 & 0.2442 & 0.6853 & 0.1198 & 27.98 \\
$\checkmark$ & $\times$ & $\times$ & $\times$ & CF-Font & 0.07394 & 0.2354 & 0.7007 & 0.1182 & 26.51 \\
\bottomrule
\end{tabular}
\label{tab:comparison}
\end{table*}

\section{Limitations and Future Work}

\begin{itemize}
    \item \textbf{Handwriting Input Potential:} Section~\ref{Made-Up} highlights the ability of our model to generate styles for made-up characters, showing our model's potential for handwriting input. However, due to time constraints, large-scale investigation was not conducted.
    
    \item \textbf{Expanding Writing Systems:} The model demonstrates strong performance across alphabetic-like scripts (e.g., Japanese Hiragana and English) and logographic characters (e.g., Kanji). Exploring its generalizability to other writing systems, such as Arabic or historical scripts like Linear A, would be an intriguing direction.
    
    \item \textbf{Few-Shot Generation:} Our model could easily extend to a few-shot setting by invoking the style encoder multiple times and averaging the style representation. While this approach has not been tested, investigating the model's performance in a few-shot scenario is an exciting avenue for future research.
\end{itemize}

\clearpage
{
    \small
    \bibliographystyle{ieeenat_fullname}
    \bibliography{main, second}

\begin{thebibliography}{26}
\providecommand{\natexlab}[1]{#1}
\providecommand{\url}[1]{\texttt{#1}}
\expandafter\ifx\csname urlstyle\endcsname\relax
  \providecommand{\doi}[1]{doi: #1}\else
  \providecommand{\doi}{doi: \begingroup \urlstyle{rm}\Url}\fi

\bibitem[noa({\natexlab{a}})]{noauthor_51font_nodate}
{51Font}, {\natexlab{a}}.

\bibitem[noa({\natexlab{b}})]{noauthor_browse_nodate}
Browse {Fonts}, {\natexlab{b}}.

\bibitem[Azadi et~al.(2017)Azadi, Fisher, Kim, Wang, Shechtman, and Darrell]{azadi_multi-content_2017}
Samaneh Azadi, Matthew Fisher, Vladimir Kim, Zhaowen Wang, Eli Shechtman, and Trevor Darrell.
\newblock Multi-{Content} {GAN} for {Few}-{Shot} {Font} {Style} {Transfer}, 2017.
\newblock arXiv:1712.00516 [cs].

\bibitem[Bao et~al.(2022)Bao, Dong, Piao, and Wei]{bao_beit_2022}
Hangbo Bao, Li Dong, Songhao Piao, and Furu Wei.
\newblock {BEiT}: {BERT} {Pre}-{Training} of {Image} {Transformers}, 2022.
\newblock arXiv:2106.08254.

\bibitem[Borgeaud et~al.(2022)Borgeaud, Mensch, Hoffmann, Cai, Rutherford, Millican, Driessche, Lespiau, Damoc, Clark, Casas, Guy, Menick, Ring, Hennigan, Huang, Maggiore, Jones, Cassirer, Brock, Paganini, Irving, Vinyals, Osindero, Simonyan, Rae, Elsen, and Sifre]{borgeaud_improving_2022}
Sebastian Borgeaud, Arthur Mensch, Jordan Hoffmann, Trevor Cai, Eliza Rutherford, Katie Millican, George van~den Driessche, Jean-Baptiste Lespiau, Bogdan Damoc, Aidan Clark, Diego de~Las Casas, Aurelia Guy, Jacob Menick, Roman Ring, Tom Hennigan, Saffron Huang, Loren Maggiore, Chris Jones, Albin Cassirer, Andy Brock, Michela Paganini, Geoffrey Irving, Oriol Vinyals, Simon Osindero, Karen Simonyan, Jack~W. Rae, Erich Elsen, and Laurent Sifre.
\newblock Improving language models by retrieving from trillions of tokens, 2022.
\newblock arXiv:2112.04426.

\bibitem[Dosovitskiy et~al.(2021)Dosovitskiy, Beyer, Kolesnikov, Weissenborn, Zhai, Unterthiner, Dehghani, Minderer, Heigold, Gelly, Uszkoreit, and Houlsby]{dosovitskiy_image_2021}
Alexey Dosovitskiy, Lucas Beyer, Alexander Kolesnikov, Dirk Weissenborn, Xiaohua Zhai, Thomas Unterthiner, Mostafa Dehghani, Matthias Minderer, Georg Heigold, Sylvain Gelly, Jakob Uszkoreit, and Neil Houlsby.
\newblock An {Image} is {Worth} 16x16 {Words}: {Transformers} for {Image} {Recognition} at {Scale}, 2021.
\newblock arXiv:2010.11929 [cs].

\bibitem[Fogel et~al.(2020)Fogel, Averbuch-Elor, Cohen, Mazor, and Litman]{fogel_scrabblegan_2020-1}
Sharon Fogel, Hadar Averbuch-Elor, Sarel Cohen, Shai Mazor, and Roee Litman.
\newblock {ScrabbleGAN}: {Semi}-{Supervised} {Varying} {Length} {Handwritten} {Text} {Generation}, 2020.
\newblock arXiv:2003.10557 [cs].

\bibitem[Hayashi et~al.(2019)Hayashi, Abe, and Uchida]{hayashi_glyphgan_2019}
Hideaki Hayashi, Kohtaro Abe, and Seiichi Uchida.
\newblock {GlyphGAN}: {Style}-{Consistent} {Font} {Generation} {Based} on {Generative} {Adversarial} {Networks}, 2019.
\newblock arXiv:1905.12502 [cs].

\bibitem[He et~al.(2024)He, Chen, Wang, Liu, Du, Tao, and Yu]{he_diff-font_2024}
Haibin He, Xinyuan Chen, Chaoyue Wang, Juhua Liu, Bo Du, Dacheng Tao, and Qiao Yu.
\newblock Diff-{Font}: {Diffusion} {Model} for {Robust} {One}-{Shot} {Font} {Generation}.
\newblock \emph{International Journal of Computer Vision}, 2024.

\bibitem[He et~al.(2021)He, Chen, Xie, Li, Dollár, and Girshick]{he_masked_2021}
Kaiming He, Xinlei Chen, Saining Xie, Yanghao Li, Piotr Dollár, and Ross Girshick.
\newblock Masked {Autoencoders} {Are} {Scalable} {Vision} {Learners}, 2021.
\newblock arXiv:2111.06377 version: 2.

\bibitem[Lee et~al.(2022)Lee, Baek, and Choi]{lee_arbitrary_2022}
Jeong-Sik Lee, Rock-Hyun Baek, and Hyun-Chul Choi.
\newblock Arbitrary {Font} {Generation} by {Encoder} {Learning} of {Disentangled} {Features}.
\newblock \emph{Sensors}, 22\penalty0 (6):\penalty0 2374, 2022.
\newblock Number: 6 Publisher: Multidisciplinary Digital Publishing Institute.

\bibitem[Li et~al.(2021)Li, Taniguchi, Lu, and Konomi]{li_few-shot_2021}
Chenhao Li, Yuta Taniguchi, Min Lu, and Shin’ichi Konomi.
\newblock Few-shot {Font} {Style} {Transfer} between {Different} {Languages}.
\newblock In \emph{2021 {IEEE} {Winter} {Conference} on {Applications} of {Computer} {Vision} ({WACV})}, pages 433--442, 2021.
\newblock ISSN: 2642-9381.

\bibitem[Liu and Lian(2022)]{liu_fonttransformer_2022}
Yitian Liu and Zhouhui Lian.
\newblock {FontTransformer}: {Few}-shot {High}-resolution {Chinese} {Glyph} {Image} {Synthesis} via {Stacked} {Transformers}, 2022.
\newblock arXiv:2210.06301 [cs].

\bibitem[Pippi et~al.(2023)Pippi, Cascianelli, and Cucchiara]{pippi_handwritten_2023}
Vittorio Pippi, Silvia Cascianelli, and Rita Cucchiara.
\newblock Handwritten {Text} {Generation} from {Visual} {Archetypes}, 2023.
\newblock arXiv:2303.15269 [cs].

\bibitem[Ryali et~al.(2023)Ryali, Hu, Bolya, Wei, Fan, Huang, Aggarwal, Chowdhury, Poursaeed, Hoffman, Malik, Li, and Feichtenhofer]{ryali_hiera_2023}
Chaitanya Ryali, Yuan-Ting Hu, Daniel Bolya, Chen Wei, Haoqi Fan, Po-Yao Huang, Vaibhav Aggarwal, Arkabandhu Chowdhury, Omid Poursaeed, Judy Hoffman, Jitendra Malik, Yanghao Li, and Christoph Feichtenhofer.
\newblock Hiera: {A} {Hierarchical} {Vision} {Transformer} without the {Bells}-and-{Whistles}, 2023.
\newblock arXiv:2306.00989.

\bibitem[Sun et~al.(2018)Sun, Ren, Li, Su, and Zhu]{sun_learning_2018}
Danyang Sun, Tongzheng Ren, Chongxun Li, Hang Su, and Jun Zhu.
\newblock Learning to {Write} {Stylized} {Chinese} {Characters} by {Reading} a {Handful} of {Examples}, 2018.
\newblock arXiv:1712.06424 [cs, stat].

\bibitem[Tang et~al.(2022)Tang, Cai, Liu, Hong, Gong, Fan, Han, Liu, Ding, and Wang]{tang_few-shot_2022}
Licheng Tang, Yiyang Cai, Jiaming Liu, Zhibin Hong, Mingming Gong, Minhu Fan, Junyu Han, Jingtuo Liu, Errui Ding, and Jingdong Wang.
\newblock Few-{Shot} {Font} {Generation} by {Learning} {Fine}-{Grained} {Local} {Styles}, 2022.
\newblock arXiv:2205.09965 [cs].

\bibitem[Tian(2024{\natexlab{a}})]{tian_kaonashi-tycrewrite_2024}
Yuchen Tian.
\newblock kaonashi-tyc/{Rewrite}, 2024{\natexlab{a}}.
\newblock original-date: 2016-10-26T03:11:46Z.

\bibitem[Tian(2024{\natexlab{b}})]{tian_kaonashi-tyczi2zi_2024}
Yuchen Tian.
\newblock kaonashi-tyc/zi2zi, 2024{\natexlab{b}}.
\newblock original-date: 2017-02-17T23:18:04Z.

\bibitem[Vaswani et~al.(2023)Vaswani, Shazeer, Parmar, Uszkoreit, Jones, Gomez, Kaiser, and Polosukhin]{vaswani_attention_2023}
Ashish Vaswani, Noam Shazeer, Niki Parmar, Jakob Uszkoreit, Llion Jones, Aidan~N. Gomez, Lukasz Kaiser, and Illia Polosukhin.
\newblock Attention {Is} {All} {You} {Need}, 2023.
\newblock arXiv:1706.03762 [cs].

\bibitem[Wen et~al.(2021)Wen, Li, Han, and Yuan]{wen_zigan_2021}
Qi Wen, Shuang Li, Bingfeng Han, and Yi Yuan.
\newblock {ZiGAN}: {Fine}-grained {Chinese} {Calligraphy} {Font} {Generation} via a {Few}-shot {Style} {Transfer} {Approach}.
\newblock In \emph{Proceedings of the 29th {ACM} {International} {Conference} on {Multimedia}}, pages 621--629, 2021.
\newblock arXiv:2108.03596 [cs].

\bibitem[Xie et~al.(2021)Xie, Chen, Sun, and Lu]{xie_dg-font_2021}
Yangchen Xie, Xinyuan Chen, Li Sun, and Yue Lu.
\newblock {DG}-{Font}: {Deformable} {Generative} {Networks} for {Unsupervised} {Font} {Generation}, 2021.
\newblock arXiv:2104.03064 [cs].

\bibitem[Xue et~al.(2023)Xue, Wang, Bu, Li, and Zhang]{xue_metascript_2023}
Xiangyuan Xue, Kailing Wang, Jiazi Bu, Qirui Li, and Zhiyuan Zhang.
\newblock {MetaScript}: {Few}-{Shot} {Handwritten} {Chinese} {Content} {Generation} via {Generative} {Adversarial} {Networks}, 2023.
\newblock arXiv:2312.16251 [cs].

\bibitem[Yang et~al.(2023)Yang, Peng, Kong, Zhang, Yao, and Jin]{yang_fontdiffuser_2023}
Zhenhua Yang, Dezhi Peng, Yuxin Kong, Yuyi Zhang, Cong Yao, and Lianwen Jin.
\newblock {FontDiffuser}: {One}-{Shot} {Font} {Generation} via {Denoising} {Diffusion} with {Multi}-{Scale} {Content} {Aggregation} and {Style} {Contrastive} {Learning}, 2023.
\newblock arXiv:2312.12142 [cs].

\bibitem[Yu(2022)]{yu_few_2022}
Yong Yu.
\newblock Few {Shot} {POP} {Chinese} {Font} {Style} {Transfer} using {CycleGAN}.
\newblock \emph{Journal of Physics: Conference Series}, 2171\penalty0 (1):\penalty0 012031, 2022.
\newblock Publisher: IOP Publishing.

\bibitem[Zhang et~al.(2018)Zhang, Zhang, Cai, and Chang]{zhang_separating_2018}
Yexun Zhang, Ya Zhang, Wenbin Cai, and Jie Chang.
\newblock Separating {Style} and {Content} for {Generalized} {Style} {Transfer}, 2018.
\newblock arXiv:1711.06454 [cs].

\end{thebibliography}
}


\end{document}